\documentclass{article}
 
\usepackage{arxiv}
 
\usepackage[utf8]{inputenc}
\usepackage[T1]{fontenc}
\usepackage{textcomp}        
\usepackage{hyperref}
\usepackage{url}
\usepackage{booktabs}
\usepackage{amsmath,amssymb,amsfonts,array}
\usepackage{nicefrac}
\usepackage{microtype}
\usepackage{graphicx}
\usepackage{natbib}
\usepackage{doi}
 
\usepackage[ruled,vlined]{algorithm2e}
\usepackage{subcaption}
\usepackage{float}
\usepackage{makecell}

\usepackage{xcolor}
\usepackage{footmisc}
\usepackage{fancyvrb}        
 
\newcommand{\Var}{\mathrm{Var}}
 
\newcommand{\code}[1]{\texttt{#1}}

\let\pkg=\strong
\newcommand{\CRANpkg}[1]{\href{https://CRAN.R-project.org/package=#1}{\pkg{#1}}}

\shortcites{tensorflow2015-whitepaper}
 
\title{neuralGAM: An R Package for Fitting Generalized Additive Neural Networks}
 
\author{
  \href{https://orcid.org/0000-0002-8041-6860}{\includegraphics[scale=0.06]{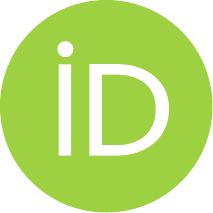}\hspace{1mm}Ines Ortega-Fernandez} \\
  Galician Research and Development Center in Advanced Telecommunications (Gradiant), Vigo (Spain)\\
  \texttt{iortega@gradiant.org} \\
  \And
  \href{https://orcid.org/0000-0003-4284-6509}{\includegraphics[scale=0.06]{orcid.pdf}\hspace{1mm}Marta Sestelo} \\
  Galician Centre for Mathematical Research and Technology (CITMAga), Santiago de Compostela (Spain)\\
  Universidade de Vigo, Department of Statistics and O.R. \& SiDOR Group, 36310 Vigo (Spain) \\
  \texttt{sestelo@uvigo.gal} \\
}
 
\renewcommand{\undertitle}{Accepted version}
\renewcommand{\headeright}{The R Journal}
\renewcommand{\shorttitle}{neuralGAM}
 
\date{%
  \textit{The R Journal} (2026), \textbf{18}(1), 253--276.\\
  \href{https://doi.org/10.32614/RJ-2026-016}{doi:10.32614/RJ-2026-016}%
}
 
\hypersetup{
  pdftitle={neuralGAM: An R Package for Fitting Generalized Additive Neural Networks},
  pdfsubject={The R Journal 18(1):253-276, doi:10.32614/RJ-2026-016},
  pdfauthor={Ines Ortega-Fernandez, Marta Sestelo},
  pdfkeywords={Explainable AI, Generalized additive models, Interpretable deep learning},
}
 
\begin{document}
\maketitle
 
\begin{center}\small
This is the author's accepted version. The version of record is published in
\textit{The R Journal} and is available at
\href{https://rjournal.github.io/articles/RJ-2026-016/}%
     {rjournal.github.io/articles/RJ-2026-016}.\\[2pt]
\textit{Please cite as:} Ortega-Fernandez, I. and Sestelo, M. (2026).
neuralGAM: An R Package for Fitting Generalized Additive Neural Networks.
\textit{The R Journal}, 18(1), 253--276.
\href{https://doi.org/10.32614/RJ-2026-016}{doi:10.32614/RJ-2026-016}.
\end{center}
 
\begin{abstract}
Nowadays, neural networks are considered one of the most effective methods for various tasks such as anomaly detection, computer-aided disease detection, or natural language processing. However, these networks suffer from the ``black-box'' problem which makes it difficult to understand how they make decisions. In order to solve this issue, an \textsf{R} package called \CRANpkg{neuralGAM} is introduced. This package implements a Neural Network topology based on Generalized Additive Models, allowing users to fit an independent Neural Network to estimate the contribution of each feature to the output variable, yielding a highly accurate and interpretable deep learning model. The \CRANpkg{neuralGAM} package provides a flexible framework for training Generalized Additive Neural Networks, which does not impose any restrictions on the per-term neural network architecture, while the overall model remains additive. We illustrate the use of the \CRANpkg{neuralGAM} package in both synthetic and real data examples.
\end{abstract}
 

\section{Introduction}

Neural networks (NN) are currently among the most widely used predictive modeling techniques, demonstrating superior performance across a broad range of tasks. Despite their success, a well-known shortcoming of neural networks is that they often function as ``black-box'' models. In most practical applications, it can be challenging to understand precisely \emph{how} the network processes information and reaches a particular prediction \citep{szegedy2013intriguing}. In recent years, there has been a growing research focus on enhancing trust in AI systems by improving their interpretability. Broadly, interpretability methods can be categorized into \emph{post-hoc} and \emph{ante-hoc} approaches \citep{dovsilovic2018explainable}. \emph{Post-hoc} methods aim to explain a trained black-box model (like a typical Neural Network) using an external, intrinsically interpretable (``white-box'') model to approximate or interpret the black-box decisions. In contrast, \emph{ante-hoc} methods aim to build and train inherently interpretable (``white-box'') models that balance high predictive accuracy with transparency. 

Following this \emph{ante-hoc} approach, several approaches in the literature have been proposed to combine the interpretability of GAMs with the flexibility of neural networks, yielding the family of Generalized Additive Neural Networks (GANNs). Early approaches, such as the model by \citet{GNAM}, used univariate multi-layer perceptrons without backpropagation and required manual tuning. Subsequent approaches, like that of \citet{Bras-Geraldes2019}, improved flexibility with parametric link functions and bootstrap-based confidence intervals but remained constrained to shallow architectures. More recent methods implemented in Python take advantage of deep learning techniques: Neural Additive Models (NAMs) \citep{NEURIPS2021_251bd044} train feature-specific subnetworks jointly with backpropagation but can suffer from overfitting due to sharp activation functions. GAMI-Net \citep{yang2021gami} extends this by allowing pairwise interactions and incorporating sparsity and marginal clarity constraints, but with an increased computational cost. Finally, IGANN \citep{kraus2023interpretable} introduces a GANN based on gradient boosting techniques with sparsified shallow networks, assuming initially linear effects and avoiding deep architectures.

The development of the \CRANpkg{neuralGAM} package has been motivated by recent contributions in the development of GANNs, in particular, the method proposed by \citet{neuralGAM} to train a GANN using independent neural networks using the local scoring (LS) and backfitting (BF) algorithms. 

Specifically, \CRANpkg{neuralGAM} trains an ensemble of independent neural networks (one per feature) to learn each feature’s contribution to the response. By enforcing an additive decomposition via backfitting and local scoring, it prioritizes interpretability over the flexibility of unconstrained deep networks: the aim is not to match their full expressive power but to strike a practical balance, delivering competitive predictive performance while maintaining the GAM-style additivity that supports clear feature-level interpretation.

Regarding other R packages that combine additive models with neural networks, the \CRANpkg{deepregression} package \citep{deepregression} implements distributional regression models for mean, scale, and shape parameters using deep neural networks, extending the GAM framework by embedding spline terms directly into a neural network architecture. Smooth functions are pre-processed using \CRANpkg{mgcv} \citep{mgcv}, which generates the spline basis and penalty matrices; these are then represented as structured layers within a deep neural network model and regularized through a TensorFlow’s loss function. 

The \CRANpkg{bamlss} package\citep{bamlss}, in contrast, provides a more general Bayesian framework for additive models that can incorporate neural network terms. Estimation in this context is performed through Bayesian sampling or variational inference, which naturally provides posterior uncertainty quantification at the expense of higher computational costs. These packages illustrate complementary methodological directions: \CRANpkg{deepregression} extends GAMs toward distributional regression with deep learning components, \CRANpkg{bamlss} offers a Bayesian approach with uncertainty quantification, and \CRANpkg{neuralGAM} provides a dedicated white-box GANN implementation through independent deep neural networks. In this regard, \CRANpkg{neuralGAM} complements existing \textsf{R} software and, to the best of our knowledge, provides the only available GANN implementation in \textsf{R} that leverages fully independent deep Neural Networks per feature and estimates the additive model through local scoring and backfitting. This design offers both interpretability and the characteristic deep learning flexibility within the \textsf{R} ecosystem.

At last, beyond GANN-specific packages, several \textsf{R} packages provide model-agnostic tools for explainability. The \CRANpkg{NeuralNetTools} package \citep{beck2018neuralnettools}, for instance, offers visualizations and sensitivity analyses tailored to neural networks. More general frameworks such as \CRANpkg{iml} \citep{iml} and \CRANpkg{DALEX} \citep{dalex} implement techniques such as feature importance, partial dependence, and Shapley values, supporting both local and global interpretability. While \CRANpkg{iml} and \CRANpkg{DALEX} are model-agnostic, \CRANpkg{NeuralNetTools} is specifically designed for neural networks.

In this paper, we demostrate  how neural networks can be used to fit Generalized Additive Models using the \CRANpkg{neuralGAM} package. This is illustrated through a simulated scenario with a Gaussian response, as well as a real-life application related to flight delay prediction based on weather and flight conditions' data. 

The remainder of the paper is structured as follows: in Section~\ref{sec:models} we briefly review the estimation procedures and explain the use of the main functions and methods of \CRANpkg{neuralGAM}; Section~\ref{sec:illustrations} gives an illustration of the practical application of the package using simulated and real data; and finally, Section \ref{sec:summary} concludes with a discussion and possible future extensions of the package. 

\section{Models and software} \label{sec:models}

\subsection{An overview of the methodology}

The most common way to model the relationship between the response variable and the covariates is by using the multiple linear regression model, where the response variable ($Y$) is assumed to be normally distributed, and the covariates ($X_j$, $j =1, \ldots, p$) are assumed to have a linear effect on the response. However, the response variable may not be normally distributed and, in such cases, Generalized Linear Models \citep{nelder1972generalized} allow the use of other distribution families, e.g. Binomial, Poisson, etc. Additionally, in some cases, the assumption of linearity in the effects of covariates can be too restrictive and not supported by the available data. In this setting, nonparametric regression techniques emerge, allowing us to model this dependence between the response and the covariates without specifying in advance the function that links them. This leads to the Generalized Additive Models \citep{hastie1990tibshirani} defined by
\begin{equation}
   E[Y\mid \textbf{X}] = m(\textbf{X}) = h^{-1}(\alpha + \sum_{j=1}^p f_j(X_j)),
\label{GAM}
\end{equation}
\noindent where $h(\cdot)$ is a monotonic known function (the link function) and $f_1, ..., f_p$ are smooth and unknown functions. To ensure the identifiability of the model (the ability to uniquely determine the parameters of the model based on the observed data), a constant denoted by $\alpha$ is introduced in the model and the partial functions must satisfy the condition $E[f_j(X_j)] = 0$ where $j = 1, \dots, p$. This implies that $E[Y] = \alpha$, a crucial condition to ensure that the model's predictions remain unchanged even if we add a constant value to $f_1$ while subtracting the same constant value from $f_2$ \citep{hastie1990tibshirani}.

As we mentioned, GAMs are widely used for their ability to capture complex relationships between covariates and response variables without requiring the shape of these relationships to be specified in advance. This flexibility makes GAMs a powerful and effective class of regression models. They also offer the advantage of being fully nonparametric while allowing certain covariates—such as categorical features—to enter the model linearly, resulting in a semi-parametric framework. Thanks to this versatility, GAMs are particularly well-suited for tackling real-world problems.

\begin{table}[ht!]
\centering
\resizebox{\textwidth}{!}{%
\begin{tabular}{ccccc}
\hline
\textbf{Distribution} & \textbf{Link}  & $Z_i$ & $W_i$ & $DEV_i(Y_i,\hat\mu_i)$ \\
\midrule
Gaussian & identity & $Y_i$ & $1$ &  $(Y_i - \hat{\mu}_i)^2$\\
\midrule
Binomial $(s, \mu)$ & logit & $\eta_i + (Y_i-\mu_i)s\mu_i(1-\mu_i)$ & $s\mu_i(1-\mu_i)$ & $-2(Y_i\log\hat{\mu}_i + (1-Y_i)\log(1-\hat{\mu}_i)$\\
\midrule
Poisson & log & $\eta_i + (Y_i-\mu_i)/\mu_i$ & $\mu_i$ &$Y_i\log\frac{Y_i}{\hat{\mu}_i} - (Y_i - \hat{\mu}_i)$ \\
\bottomrule
\end{tabular}
}
\caption{Adjusted dependent variable $Z_i$, weights $W_i$, and deviance $DEV_i(Y_i,\hat\mu_i)$, at the local scoring algorithm for Gaussian, binomial and Poisson distributions \citep{hastie1990tibshirani}.}
\label{dist-link}
\end{table}

To fit the previous model in (\ref{GAM}), from an independent random sample $\{\boldsymbol{X}_i, Y_i\}_{i=1}^n$, where each $\boldsymbol{X}_i = (X_{i1}, \ldots, X_{ip})$ is a realization of the $p-$dimensional random vector 
$\mathbf{X} = (X_1, \ldots, X_p)$, we use a combination of the local scoring algorithm (see Algorithm \ref{local-scoring}) and the backfitting algorithm (see Algorithm \ref{backfitting}).

In practice, some covariates may enter the model linearly while others are represented by smooth functions. In this case, for the model in (\ref{GAM}), each $f_j$ may either be a smooth, nonparametric function fitted using a neural network or a parametric linear effect of the form $f_j(x_j)=x_j^\top \beta_j$, leading to a semiparametric model.

Particularly, the estimation of the additive predictor $\eta = \alpha + \sum_{j=1}^{p} f_j (\cdot)$ is obtained by fitting a weighted additive model where each $f_j$ is either a smooth function (estimated by a neural network) or a linear effect, yielding a Generalized Additive Neural Network. For each smooth $f_j$, the backfitting algorithm iteratively estimates the effect of each covariate $X_1, \ldots, X_p$, with weights $W_i$, on the adjusted dependent variable $Z_i$, updated at each step of the local scoring algorithm. 

\begin{algorithm}[ht!]
\SetAlgoLined
\caption{Local scoring algorithm}
\label{local-scoring}
\textbf{Initialise} $\hat \alpha = h(\overline{Y_i}), \hat f_j^0 (\cdot) = 0 \, \forall \, j, \, l = 0 $ \BlankLine
\For {$l \leftarrow l + 1$}{
    \BlankLine
    Construct an adjusted dependent variable $Z_i$ according to Table \ref{dist-link} with
    \[
    \hat \eta_i^{\,l-1} = \hat \alpha + \sum_{j=1}^p \hat f_j^{\,l-1}(X_{ij}), 
    \qquad \hat \mu_i^{\,l-1} = h^{-1}(\hat \eta_i^{\,l-1}).
    \]
    \BlankLine
    Form the weights $W_i$ according to Table \ref{dist-link}. 
    \BlankLine
    \textbf{Update the linear component (if any):} fit a weighted least squares model of $Z_i$ on the linear covariates.
    Update $\hat \alpha$ to be the fitted intercept from the parametric component, and update $\hat f_j^l$ for the linear terms accordingly.
    \BlankLine
    \textbf{Update the non-linear component:} fit a weighted additive model to $Z_i$ using the backfitting algorithm (Algorithm \ref{backfitting}) 
    with weights $W_i$ to obtain the estimated functions $\hat f_j^l$, additive predictor $\hat \eta_i^l$, and fitted values $\hat{\mu}_i^l$.}
    \BlankLine
\textbf{Compute the convergence criterion} $\text{DEV} < \delta$, with DEV as in Table \ref{dist-link}. \BlankLine
\textbf{Until} the convergence criterion is satisfied \BlankLine
\end{algorithm}

Note that if the link function is the identity and the error distribution is Gaussian,  then $Z_i = Y_i$ and the weights do not change, thus the procedure is simply an additive fit. In any other case, the dependent variable $Z_i$ and the weights $W_i$ are updated at each iteration $l$  of the local scoring algorithm. This process is repeated until the convergence criterion is satisfied (see last step of Algorithm \ref{local-scoring}). Note that we use the deviance because it is an appropriate measure of the discrepancy between observed and fitted values. 

\begin{algorithm}[ht!]
    \SetAlgoLined
    \caption{Backfitting Algorithm with neural networks}
    \label{backfitting}
        \textbf{Initialize} $\hat f_j^0(\cdot) = 0 \, \forall \, j, \, m = 0 $
        
        \textbf{Iterate} $m \gets m+1$
            \BlankLine
            \For {$j \leftarrow 1, p$}{
                \BlankLine
                Compute partial residuals 
                $R_{ij} = Z_i - \hat \alpha - \sum_{k \neq j} \hat f_k^{\,m-1}(X_{ik})$ \BlankLine
                For each smooth term, fit a Neural Network with $R_{ij}$ and $X_{ij}$ to obtain $\hat f_j^m$ \BlankLine
                Replace $\hat f_j^m$ with its centered version $\hat f_j^m(\cdot) - \frac{\sum_{i=1}^{n} \hat f_j^m(X_{ij})}{n}$ \BlankLine
                \BlankLine
                }
            \textbf{Until}
            \[
            \frac{ \sum_{j=1}^p \sum_{i=1}^n \bigl(\hat f_j^{\,m}(X_{ij}) - \hat f_j^{\,m-1}(X_{ij})\bigr)^2}
                 { \sum_{j=1}^p \sum_{i=1}^n \bigl(\hat f_j^{\,m-1}(X_{ij})\bigr)^2}
            < \epsilon
            \] \BlankLine
\end{algorithm}

Particularly, given the fitted mean response $\hat{\mu}_i = \hat E[Y_i \mid \textbf{X}_i]$, the deviance is defined as

\[DEV = \frac{\sum_{i=1}^n DEV_i(Y_i, \hat \mu_i^{l-1}) - DEV_i(Y_i, \hat \mu_i^{l})}{\sum_{j=1}^n DEV_i(Y_i, \hat \mu_i^{l-1})},
\]

\noindent with $\text{DEV}_i$ depending on the link (see Table \ref{dist-link}). Several approaches have been described in the literature to estimate the regression model in (\ref{GAM}), such as Bayesian approaches \citep{lang2004bayesian}, local polynomial kernel smoothers \citep{wand1994kernel, copeland1997local} or regression splines \citep{de1978practical}. Unlike these quite common techniques, we propose to use independent neural networks ---which are universal function estimators \citep{hornik1989multilayer}--- to learn the contribution of each covariate to the dependent variable, i.e., at each iteration of the backfitting algorithm, we train a feed-forward Neural Network for each covariate $X_j$ with the entire set of training data for one epoch. The fits are improved at each epoch as the learned adjusted dependent variable $Z_i$ approaches $Y_i$ at each iteration of the local scoring algorithm. Algorithms~\ref{local-scoring} and \ref{backfitting} summarize the core idea.

\subsubsection{Uncertainty estimation}
\label{sec:uncertainty}

To quantify epistemic uncertainty, we leverage Monte Carlo (MC) Dropout \citep{gal2016dropout}, a widely used technique for estimating model uncertainty in neural networks by interpreting dropout as a Bayesian approximation. In \CRANpkg{neuralGAM}, dropout layers are inserted into the neural networks used to estimate the smooth functions $f_j$. During training, dropout layers randomly ``drop'' (set to zero) a fraction of the units in each layer of the network, acting as a regularizer to prevent overfitting. At inference time, rather than being deactivated as in standard practice, dropout remains active to generate stochasticity. The network is then evaluated repeatedly through $B$ stochastic forward passes, producing a distribution of predictions that can be used to quantify epistemic uncertainty for each individual term and for the additive predictor in the link and response scales.

\paragraph{Per-term uncertainty.}

For each smooth term $f_j$, the use of MC Dropout during inference yields $B$ draws
\[
\hat f_j^{(b)}(x_j)_{b=1}^B.
\] 

To guarantee the identifiability of the model, these smooth dropout predictions are re-centered by subtracting each term’s training-time centering constant, so that for every $j$
\[\frac{1}{n}\sum_{i=1}^n \hat f_j^{(b)}(x_{ij}) = 0.\]

The epistemic variance of each nonparametric component is therefore defined as 
\[
\widehat{\Var}_{\text{np}}\bigl[f_j(x_j)\bigr] = \widehat{\Var}_b\bigl(\hat f_j^{(b)}(x_j)\bigr),
\]
where $\hat f_j^{(b)}(x_j)$ denotes the prediction of the $j$-th component in the $b$-th simulation or dropout pass, and $\widehat{\Var}_b$ indicates the empirical variance over these $B$ realizations.

For models including a linear (parametric) component $f_j(x_j)= x_j^\top \beta_j $, the uncertainty is obtained from the fitted linear model (using weighted least squares) via the standard errors of the predicted values, $\widehat{SE}_p(f_j(x_j))$. In practice, we use the standard errors reported by the \texttt{stats::predict(..., se.fit=TRUE)} function in \textsf{R}, which are based on the variance--covariance matrix of the estimated coefficients $\hat\beta$. These quantify the sampling variability of the estimated coefficients $\hat\beta$ under the classical linear model assumptions. 

The corresponding epistemic variance for the parametric component is then defined as: 
\[
\widehat{\Var}_p(f_j(x_j)) = \left[ \widehat{SE}_p(\hat f_j(x_j)) \right]^2
\]

To provide an unified notation, we define the epistemic variance of any model component $f_j(x_j)$ as

\[
\widehat{\Var}_{\mathrm{epi}}\!\bigl[f_j(x_j)\bigr] =
\begin{cases}
\widehat{\Var}_{\mathrm{np}}\!\bigl[f_j(x_j)\bigr], & \text{if $f_j$ is a smooth term},\\[6pt]
\widehat{\Var}_{\mathrm{p}}\!\bigl[f_j(x_j)\bigr], & \text{if $f_j$ is a linear term.}
\end{cases}
\]

Since both the dropout-based and the standard error-based variances quantify epistemic uncertainty in their respective components, per-term confidence intervals at level $1-\alpha$ can be expressed in a unified form as
\[
\hat f_j(x_j)\;\pm\;
z_{1-\alpha/2}\,
\sqrt{\widehat{\Var}_{\mathrm{epi}}\!\bigl[f_j(x_j)\bigr]}.
\]

\paragraph{Uncertainty for the additive predictor on the link scale.}

Moving from individual terms to the full model, to obtain uncertainty bands for the full additive predictor
$\eta(\mathbf{x})$ on the link scale, we combine the uncertainty from the nonparametric and parametric components. 

For covariates fitted using a neural network, nonparametric epistemic variability is captured by evaluating all smooth networks jointly under dropout. At each dropout pass $b$, define the smooth-only predictor
\[
\eta_{np}^{(b)}(\mathbf{x}) = \sum_{j\in\mathcal{S}} \hat f_j^{(b)}(x_j),
\]
where $\mathcal{S}$ denotes the set of smooth terms, so that the dropout variance across the $B$ passes captures epistemic uncertainty and cross-term covariance among smooths:

\[
\widehat{\Var}_{np} \left[{\eta(\mathbf{x})} \right] = \Var\!\bigl\{\eta_{np}^{(b)}(\mathbf{x})\bigr\},
\]
where $\widehat{\Var}_{np}$ denotes the empirical variance computed over the $B$ dropout realizations.

When the model includes a parametric linear component $x_j^\top \hat\beta$, its variance is obtained from the standard errors of the linear predictor: 
\[
\widehat{\Var}_p\left[\eta(\mathbf{x})\right] = \left[ \widehat{SE_p}(\hat f(\mathbf{x})) \right]^2
\]

Assuming independence between the linear and smooth parts, the total epistemic variance on the link scale is then
\[
\widehat{\Var}_{\text{epi}}\{\eta(\mathbf{x})\}
\;=\; \Var_{np}\{\eta(\mathbf{x})\} \;+\; \Var_p\{\eta(\mathbf{x})\},
\]
and the $(1-\alpha)$ confidence interval on the link scale is 
\[
\hat\eta(\mathbf{x}) \pm z_{1-\alpha/2} \sqrt{\widehat{\Var}_{\mathrm{epi}}\!\bigl[\eta(\mathbf{x})\bigr]}.
\]

\paragraph{Uncertainty for the additive predictor on the response scale.}

Finally, uncertainty on the response scale is obtained by applying the delta method. The fitted mean is related to the additive predictor through the inverse link function, $\mu = h^{-1}(\eta)$. Since uncertainty is first quantified on the link scale, we need to translate it into the scale of $\mu$. The delta method provides a simple rule: the variance of $\mu$ can be approximated by multiplying the variance of $\eta$ by the squared slope of the transformation. Formally,
\[
\widehat{\Var} \{\mu(\mathbf{x}) \} \approx \bigl((h^{-1})'(\hat\eta(\mathbf{x}))\bigr)^2 \widehat{\Var}_{epi}\{\eta(\mathbf{x})\},
\]
where $(h^{-1})'(\eta) = \tfrac{d}{d\eta}h^{-1}(\eta)$ is the derivative of the inverse link. The corresponding standard error of the fitted mean on the response scale is therefore
\[
\widehat{SE}_\mu(\mathbf{x})
= \bigl|(h^{-1})'(\hat\eta(\mathbf{x}))\bigr|\;
\sqrt{\widehat{\Var}_{\text{epi}}\{\eta(\mathbf{x})\}},
\]
and the $(1-\alpha)$ confidence interval can be obtained as
\[
\hat\mu (\mathbf{x}) \pm z_{1-\alpha/2}\,\widehat{SE}_\mu(\mathbf{x}).
\]

This procedure ensures that uncertainty bands respect the chosen link function: with the identity link (Gaussian models) the derivative is $1$ and the intervals are unchanged, while with non-identity links (such as the logit or log links), the intervals are automatically rescaled according to the applied transformation.

\subsection{Package structure and functionality}
\label{package_struc}

The \CRANpkg{neuralGAM} package introduces a new methodology for fitting Generalized Additive Models, with Gaussian, binary, or Poisson responses, based on independent neural networks. It is composed of several functions that enable users to fit the models with the methods described above. \CRANpkg{neuralGAM} relies on \CRANpkg{keras} \citep{chollet2015keras} and \CRANpkg{TensorFlow} \citep{tensorflow2015-whitepaper}, high-level APIs for the implementation of neural networks well-known for its simplicity, flexibility, and ease of use. \CRANpkg{neuralGAM} also relies on other \textsf{R} packages for visualization (\CRANpkg{ggplot2}) and numerical utilities.

\begin{table}[ht!]
\begin{small}
\begin{tabular}{p{3.5cm}p{9.6cm}}
\toprule
Function & Description \\ 
\midrule
\code{neuralGAM} & Main function to fit a \code{neuralGAM} model. Fits per-feature neural networks with user-specified architectures, and combines them using backfitting and local scoring. Supports uncertainty quantification through Monte Carlo dropout. \\
\code{summary.neuralGAM} & Method for the generic \code{summary} function, providing numerical summaries of fitted models. \\
\code{print.neuralGAM} & Method for the generic \code{print} function, showing key model components. \\
\code{plot.neuralGAM} & Visualization of smooth and linear terms using base \textsf{R} graphics. \\
\code{autoplot.neuralGAM} & Visualization of fitted components and prediction/confidence intervals using \CRANpkg{ggplot2} \citep{ggplot2}, returning a \code{ggplot} object, with support for confidence and prediction intervals. \\
\code{predict.neuralGAM} & Prediction method with options for link, response, or term-level contributions. Supports standard errors and confidence intervals. \\
\code{diagnose} & Diagnostic plots for model evaluation, including residuals and Quantile-Quantile plots. For Gaussian models, these plots diagnose symmetry, tail behavior, and mean/variance misfit. For discrete data (binomial, Poisson), randomized Quantile residuals are available, which often yield smoother QQ behavior. \\
\code{plot\_history} & Visualization of the training and validation loss history for each term-specific network, across backfitting iterations. \\
\code{install\_neuralGAM} & Creates a \code{conda} environment (installing \code{miniconda} if required) and sets up the Python requirements to run neuralGAM (TensorFlow and Keras, version 2.15). \\
\bottomrule
\end{tabular}
\caption {Summary of functions in the \CRANpkg{neuralGAM} package.}
\label{pkg:fun}
\end{small}
\end{table}

The package is designed along lines similar to those of other \textsf{R} regression packages. The functions within \CRANpkg{neuralGAM} are briefly described in Table \ref{pkg:fun}. The main function of the package is \code{neuralGAM}, which fits a Generalized Additive Neural Network to estimate the contribution of each smooth function to the response. The arguments of this function are shown in Table \ref{pkg:arg}. Note that through the argument \code{formula} users can decide to fit a model with neural networks(\code{s(x)}), linear, or factor terms (\code{x}), and using the argument \code{family} it is possible to select the conditional distribution of the response variable. So far, the user can select between Gaussian, binomial, and Poisson.

Beyond specifying network architecture (\code{num\_units}, \code{activation}, initializers, regularizers, losses, etc.), \CRANpkg{neuralGAM} implements uncertainty estimation via the \code{uncertainty\_method} argument, allowing the computation of epistemic uncertainty. Additional arguments such as \code{alpha}, \code{forward\_passes}, \code{dropout\_rate}, provide fine control over uncertainty estimation.

Cross-validation support during neural network training is implemented via the \code{validation\_split} argument. Users can also visualize how the training and validation loss evolves after each backfitting iteration using the \code{plot\_history} function.

Numerical and graphical summaries of the fitted object can be obtained by using the \code{print}, \code{summary}, \code{plot}, \code{autoplot}, \code{plot\_history} and \code{diagnose} methods implemented for \code{neuralGAM} objects. Another of these methods is available for the \code{predict} function, which takes a fitted model of the \code{neuralGAM} class and, given a new data set of values of the covariates by means of the argument \code{newdata}, produces predictions, including uncertainty estimates in the form of confidence intervals. Diagnostic checks can be performed with the \code{diagnose} function, which produces residual plots and QQ-plots under different reference distributions. At last, we provide a helper function \code{install\_neuralGAM()} to assist the user in installing the required Python dependencies in a custom \code{conda} environment. Once the dependencies are installed using \code{install\_neuralGAM()}, the user must reload the library again using \code{library(neuralGAM)} for the changes to take effect. Note that the first time the package is used after installing and setting up the dependencies, the process of loading the required Python packages might take some time.

\begin{table}[ht!]
\begin{small}
\begin{tabular}{p{3.5cm}p{9.6cm}}
\toprule
 & \code{neuralGAM()} arguments \\
\midrule
\code{formula} & An object of class ``formula'': a description of the model to be fitted. Smooth neural network terms can be included using \code{s()} while plain variables are linear/factor terms. \\
\code{data} & A data frame containing the model response variable and covariates required by the formula. Additional terms not present in the formula will be ignored. \\
\code{family} & An string specifying the distribution and link to use for fitting. By default, \code{"gaussian"}, also \code{"binomial"} and  \code{"poisson"} for logistic and poisson regression. \\
\code{num\_units} & Defines the number of units at each Dense layer of a Neural Network. If a scalar value is provided, a single hidden layer is used. If a vector or list is provided, a deep architecture is fitted with corresponding hidden units. \\ 
\code{learning\_rate} & Learning rate for the optimizer. \\ 
\code{activation} & Activation function to use on every layer of the Neural Network. Defaults to \code{"relu"}. \\ 
\code{kernel\_initializer} & Kernel initializer for the Dense layers. Defaults to Xavier initializer \code{"glorot\_normal"} \citep{pmlr-v9-glorot10a}.\\ 
\code{kernel\_regularizer} & Optional regularizer applied to the kernel weights. \\
\code{bias\_regularizer} & Optional regularizer applied to the bias vector. \\
\code{bias\_initializer} & Optional initializer for the bias vector (default \code{"zeros"}). \\
\code{activity\_regularizer} & Optional regularizer applied to the output of the layer. \\
\code{loss} & Loss function for training. Defaults to \code{"mse"}, and can be any Keras built-in (e.g., \code{mse}, \code{mae}, \code{logcosh}, etc.) or a custom function. See \ref{per-term-arch} for more details on how to define custom loss functions. \\
\code{uncertainty\_method} & Type of predictive inference: one of \code{"none"} or \code{"epistemic"}. Enables uncertainty quantification with MC-dropout (see Section~\ref{uncertainty-est-imp}).\\
\code{alpha} & Significance level for interval estimation (default \code{0.05}). \\
\code{forward\_passes} & Number of Monte Carlo forward passes when computing epistemic uncertainty. Default \code{100}. \\
\code{dropout\_rate} & Dropout probability for MC-dropout estimation. Default \code{0.1}. \\
\code{validation\_split} & Fraction of data used as validation set during training. By default \code{NULL}. \\
\code{w\_train} & Optional sample weights. \\
\code{bf\_threshold} & Convergence criterion of the backfitting algorithm. Defaults to \code{0.001}. \\
\code{ls\_threshold} & Convergence criterion of the local scoring algorithm. Defaults to \code{0.1}. \\ 
\code{max\_iter\_backfitting} & Maximum number of iterations of the backfitting algorithm. Defaults to \code{10}. \\
\code{max\_iter\_ls} & Maximum number of iterations of the local scoring algorithm. Defaults to \code{10}. \\
\code{seed} & Random number generator seed for reproducibility. \\
\code{verbose} & Verbosity mode (0 = silent, 1 = print messages). Defaults to \code{1}. \\
\code{\ldots} & Other arguments to pass on to the Adam optimizer \citep{kingma2014adam}. \\ 
\bottomrule
\end{tabular}
\caption {Arguments of \code{neuralGAM()} function.}
\label{pkg:arg}
\end{small}
\end{table}

\subsubsection{Installing and setting up the package}

Since \CRANpkg{neuralGAM} relies on \CRANpkg{keras} for fitting the underlying neural network models, the package requires a working installation of Python (version $3.10$) with \CRANpkg{keras} and its TensorFlow backend (version $2.15$). The setup can be performed directly from within \CRANpkg{neuralGAM}, ensuring that all necessary dependencies are installed before using the package. 

To ease this process, we provide the \code{install\_neuralGAM()} to automate the Python setup required to run \CRANpkg{neuralGAM}, which also creates a custom \code{conda} environment to isolate the package requirements for \CRANpkg{neuralGAM}. This ensures no conflict with previous installations or dependencies from other \textsf{R} packages.

From the user’s perspective, installation of all Python dependencies requires only a single call:

\begin{verbatim}
library(neuralGAM)
# one-time setup of Python dependencies
install_neuralGAM()
# reload neuralGAM to activate environment
library(neuralGAM)  
\end{verbatim}

After running the function (and restarting the \textsf{R} session), the user can immediately fit neural additive models with \CRANpkg{neuralGAM} without needing to manually install Python, TensorFlow, or Keras.

\subsubsection{Basic functionality}

An example of the use of \CRANpkg{neuralGAM}, illustrating its basic features, is given in the following example of code. \CRANpkg{neuralGAM} is configured to fit a Deep Neural Network with \code{64} units on each of its three deep layers (\code{num\_units}). In this example, the linear predictor is composed of a linear term in $x$ and neural network fitted terms for $z$ and $w$, and the response is assumed to follow a Gaussian distribution (\code{family}):

\begin{Verbatim}
neuralGAM(y ~ x + s(z) + s(w), data = data, num_units = c(64, 64, 64),
    family = "gaussian", learning_rate = 0.001, activation = "relu",
    kernel_initializer = "glorot_normal", loss = "mse", 
    kernel_regularizer = NULL, bias_regularizer = NULL,
    uncertainty_method = "none",
    bias_initializer = 'zeros', activity_regularizer = NULL,
    bf_threshold = 0.001, ls_threshold = 0.1, max_iter_backfitting = 10,
    max_iter_ls = 10, seed = NULL, verbose = 0, ...)
\end{Verbatim}

Other arguments of \CRANpkg{neuralGAM} will be familiar to \CRANpkg{keras} users: \code{activation} defines the activation function for the Dense layers of each fitted Neural Network (defaults to \code{relu}). \CRANpkg{neuralGAM} uses Adam \citep{kingma2014adam} as an optimizer for stochastic gradient descent, whose behaviour can be customized using the \code{loss}, \code{activation}, \code{learning\_rate}, \code{kernel\_initializer}, \code{kernel\_regularizer}, \code{bias\_regularizer}, \code{bias\_initializer}, and \code{activity\_regularizer} arguments.

The backfitting and local scoring algorithms used to fit the GAM model can be configured using the \code{bf\_threshold} and \code{ls\_threshold} arguments which adjust the convergence criterion of both algorithms, while \code{max\_iter\_backfitting} and \code{max\_iter\_ls} adjust the maximum number of iterations of each algorithm. Finally, \code{seed} specifies an optional random number generator seed for algorithms dependent on randomization, \code{verbose} sets the verbosity of the printed outputs, while the \code{\ldots} argument allows setting other arguments available for the Adam implementation in \code{keras} such as the exponential decay rate for the 1st and 2nd-moment estimates.
    
The \code{summary} method returns a summary of the fit where it is possible to observe the model architecture for both the neural networks and the linear terms (if working in a semi-parametric setting) and the per-term network configuration (units at each layer, activation function, and training parameters and settings) and network layer configuration (such as layer type or regularization and initialization configurations). When using \code{uncertainty\_method = "epistemic"}, dropout layers are included in the model architecture. These dropout layers are not only used during training for regularization, but also at prediction time (through Monte Carlo dropout) to quantify epistemic uncertainty and construct confidence intervals. At last, the \code{summary} function provides a summary of the model training history, including information about the evolution of the model fit for each covariate, allowing the monitoring of both the training loss (in this case, the Mean Squared Error \code{mse}), the deviance explained by the model, and the elapsed time on each training epoch. The training loss per-term and backfitting iteration can also be visualized using the \code{plot\_history()} function.

Creating efficient and appropriate data visualizations becomes increasingly important as the complexity of the data increases. \CRANpkg{neuralGAM} provides two different methods for plotting data: one based on \textsf{R}'s standard plotting function (\code{plot.default}) in which the \code{plot} function inherits all arguments from \code{graphics} package and can be set as usual, and one based on  \code{ggplot2} \citep{ggplot2}. With this latter plot method, the \code{autoplot} function creates a \code{ggplot} object that can be modified later by the user just using the functionality provided in the \code{ggplot2} package. The following excerpt of code shows analogous plots using the two methods implemented:

\begin{Verbatim}
plot(ngam, main = "My plot")
library(ggplot2)
autoplot(ngam, which = "terms", term = "z") + ggtitle("My plot")
\end{Verbatim}

The \code{predict} method for a fitted \code{neuralGAM} produces predictions on the link, response, or per-term scales via the \code{type} argument (default \code{"link"}). If \code{newdata} is omitted, the cached training data are used:

\begin{Verbatim}
pred_link <- predict(ngam, test, type = "link")
\end{Verbatim}

We can also obtain the prediction on the response scale:

\begin{Verbatim}
pred_resp <- predict(ngam, test, type = "response")
\end{Verbatim}

At last, a subset of per-term contributions can be selected with the \code{terms} argument when \code{type = "terms"} (the intercept is never included): 

\begin{Verbatim}
terms <- predict(ngam, newdata = test, type = "terms")
terms <- predict(ngam, newdata = test, type = "terms", terms = c("z", "w"))
\end{Verbatim}

\subsubsection{Uncertainty estimation and visualization} \label{uncertainty-est-imp}

Uncertainty estimates can be obtained directly through the \code{predict()} and \code{autoplot()} methods. For example, by setting \code{uncertainty\_method = "epistemic"} during the fitting of the model, the user can request confidence intervals (CIs) for the conditional mean based on MC-Dropout. As noted in Table~\ref{pkg:arg}, \code{alpha} is the significance level, \code{dropout\_rate} controls the Dropout probability, and \code{forward\_passes} defines the number of stochastic passes to be run:

\begin{Verbatim}
ngam <- neuralGAM(
  y ~ x + s(z) + s(w),
  data = data,
  num_units = c(64,64,64),
  uncertainty_method = "epistemic",
  alpha = 0.05, dropout_rate = 0.01, 
  forward_passes = 500
)
\end{Verbatim}

To obtain the uncertainty in predictions, the argument \code{se.fit = TRUE} enables the computation of standard errors for the fitted mean. These errors capture only the \emph{epistemic} uncertainty, estimated through the MC-Dropout procedure. When \code{type = "response"} is specified, the standard errors are mapped from the link scale to the response scale using the delta method. In addition to standard errors, the \code{predict} method also returns the lower and upper CIs at level $1-\alpha$: 

\begin{Verbatim}
# obtain predictions with confidence intervals
pred <- predict(ngam, 
                which = "response", 
                se.fit = TRUE, 
                interval = "confidence", 
                alpha = 0.05)

# visualize epistemic uncertainty
autoplot(ngam, 
        which = "terms", 
        term terms = "w", 
        interval = "confidence")
\end{Verbatim}

\subsubsection{Validation on a hold-out set}

A common concern when training neural networks is the risk of overfitting if the training hyperparameters are not properly tuned. \CRANpkg{neuralGAM} includes cross-validation and hold-out splits via the \code{validation\_split} argument, which reserves a user-specified fraction of the training data for validation purposes at each backfitting iteration. During model training, both training and validation losses are stored for every term-specific network and at every iteration, and can be visualized using the \code{plot\_history()} function. This enables the user to monitor overfitting, adjust early stopping, and tune hyperparameters (such as \code{num\_units}, \code{dropout\_rate}, or \code{learning\_rate}) in an informed way.

For more advanced evaluation using cross-validation, users can combine \CRANpkg{neuralGAM} with standard resampling approaches in \textsf{R}, such as \CRANpkg{caret} or \CRANpkg{rsample}, by looping over training/validation folds and fitting multiple \code{neuralGAM} models:

\begin{Verbatim}
if(!require("rsample")) install.packages("rsample")
library(rsample)

# Assuming dat is a data.frame with x1, x2 and response y

# 5-fold cross-validation
cv_folds <- vfold_cv(dat, v = 5)

cv_results <- lapply(cv_folds$splits, function(split) {
  train <- analysis(split)
  test  <- assessment(split)
  
  # Fit neuralGAM with a validation split
  ngam <- neuralGAM(
    y ~ s(x1) + x2,
    data = train,
    family = "gaussian",
    num_units = c(64, 64),
    validation_split = 0.2,
    uncertainty_method = "epistemic", forward_passes = 50
  )
  
  # Predict on test set
  preds <- predict(ngam, newdata = test, type = "response")
  
  # Compute RMSE for this fold
  sqrt(mean((test$y - preds)^2))
})

mean_rmse <- mean(unlist(cv_results))
print(mean_rmse)
\end{Verbatim}
    
\subsubsection{Per-term architecture and custom loss definition} \label{per-term-arch}
As previously introduced, beyond global hyperparameters, \CRANpkg{neuralGAM} allows the user to specify architectural choices individually for each smooth term. For example:
\begin{Verbatim}
y ~ s(x1, num_units = c(256,128), activation = "tanh") +
    s(x2, num_units = 64, kernel_regularizer = ::regularizer_l2(1e-4)) +
    x3
\end{Verbatim}

In this example, the smooth effect of \code{x1} is estimated with a two-layer network of sizes 256 and 128 using \code{tanh} activations, while \code{x2} is modeled with a smaller single-layer network with $L_2$ regularization. Terms not explicitly specified inherit the global defaults given in the main function call. This per-term flexibility allows customizing the training of complex covariates while keeping simpler covariates modeled with lightweight networks, which can improve both predictive performance and interpretability.

Additionally, \CRANpkg{neuralGAM} allows the user to fit the model with any custom-built loss function, in addition to the standard losses provided by \CRANpkg{keras}. The loss can be passed as a \code{keras} or \code{tensorflow} loss object, or defined directly by the user as an \textsf{R} function returning a valid tensor operation. This feature enables advanced use cases such as optimizing alternative objectives (e.g. Huber loss for robustness) or incorporating problem-specific penalties (e.g. monotonicity penalties):

\begin{Verbatim}
huber_delta <- function(delta = 0.3) {
  function(y_true, y_pred) {
    tensorflow::tf$keras$losses$Huber(delta = delta)(y_true, y_pred)
  }
}

fit_ng <- neuralGAM(
  formula = y ~ s(x1) + s(x2) + s(x3),
  data = df,
  num_units = c(32, 32),
  activation = "relu",
  learning_rate = 1e-3,
  uncertainty_method = "epistemic",
  alpha = 0.95,
  # Use custom loss
  loss = huber_delta(0.3)
)
\end{Verbatim}

\subsubsection{Diagnosis tools for model validation}

\begin{figure}[H]
    \centering
    \includegraphics[width=0.5\linewidth]{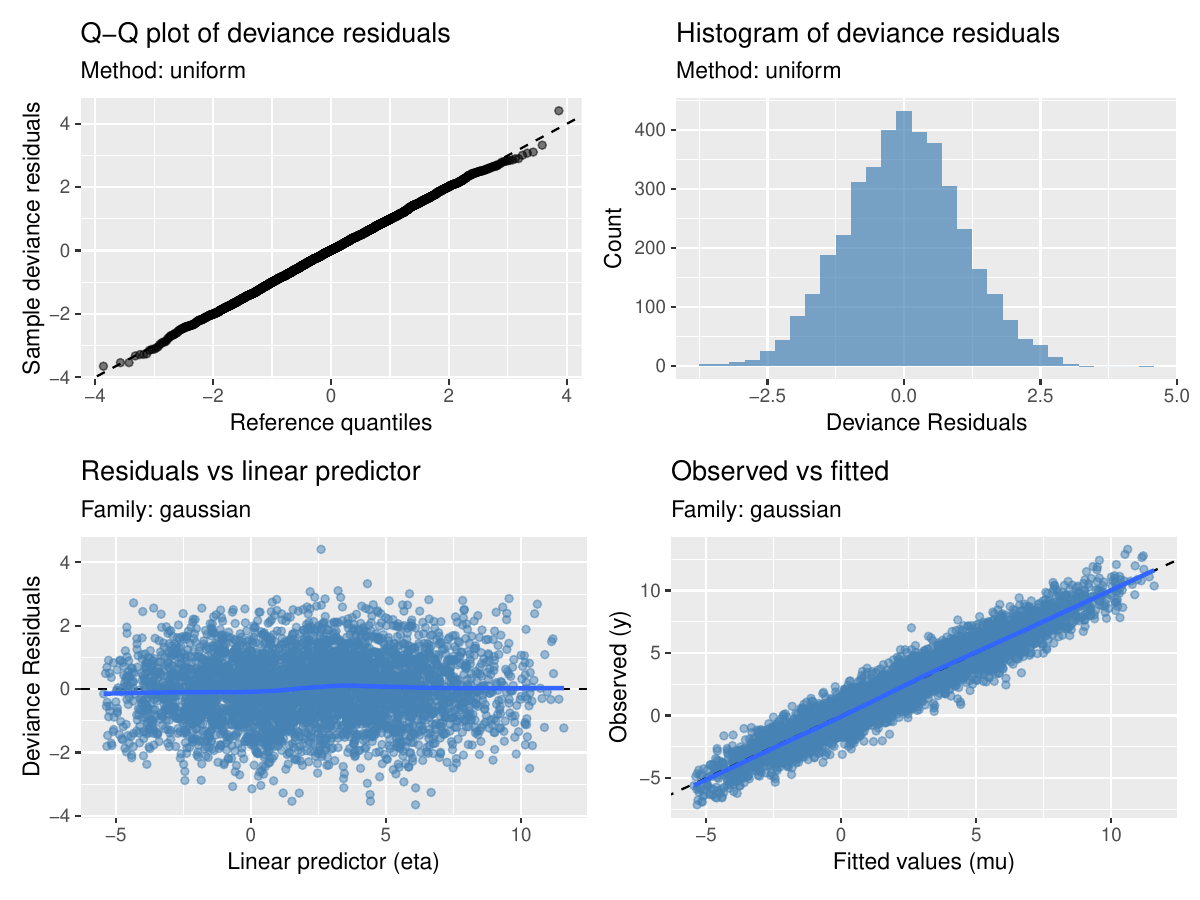}
    \caption{Example output from the \code{diagnose} function for a Gaussian model}
    \label{fig:diagnose-example}
\end{figure}

When fitting any machine learning model, it is essential to verify whether model assumptions still hold. The function \code{diagnose()} provides a compact $2\times2$ grid of diagnostic plots (see Figure~\ref{fig:diagnose-example}) summarizing key aspects of model fit: a Q--Q plot, residual histogram, residuals versus linear predictor, and observed versus fitted values. Residuals may be computed as deviance, Pearson, or quantile residuals by means of the \code{residual\_type} argument, allowing users to focus on distributional fit, or smooth normal diagnostics for discrete families. The Q--Q plot supports multiple strategies for generating theoretical quantiles by means of the \code{qq\_method} argument, including uniform and simulation-based methods that respect the fitted response distribution, as well as a normal reference option.

These four diagnostic plots provide an easy-to-use and interpretable summary to verify model assumptions in \CRANpkg{neuralGAM}, allowing users to assess whether residual distributions are consistent with model assumptions, detect potential problems such as heteroskedasticity, nonlinearity, or overdispersion, and to evaluate the adequacy of the link functions.

\section{Illustrative examples}\label{sec:illustrations}

In this section, the use of the \CRANpkg{neuralGAM} package is illustrated using both simulated and real data. We consider examples for different conditional distributions of the response variable, i.e., both Gaussian and binary. 
To contextualize the performance and interpretability of \CRANpkg{neuralGAM}, we also include a comparative analysis with the \CRANpkg{deepregression} package in both the simulated and real data scenarios, highlighting the differences in modeling flexibility, smoothness of the recovered partial functions, and predictive performance between the two approaches. In the application to real data, we also compare the predictive performance of \CRANpkg{neuralGAM} and \CRANpkg{deepregression} to a black-box neural network implemented in \CRANpkg{keras}. The complete code for the generation of the simulated data and to reproduce the results from this section is publicly available at GitHub\footnote{\url{https://github.com/inesortega/neuralGAM/blob/main/examples.R}}.

\subsection{Application to simulated data}

This subsection illustrates the capabilities of the \CRANpkg{neuralGAM} package in a controlled simulation study. We generated a dataset of size $n = 30625$, split into 80\% for training and 20\% for testing. We consider the following predictor
\begin{equation*}
   \eta = \alpha + \sum_{j=1}^3 f_j(X_j),
\end{equation*}
\noindent with
\begin{equation*}
    f_j(X_j) = 
    \begin{cases}
    X_j^2 \hspace{1.4cm} \text{ if } j=1, \\ 
    2X_j \hspace{1.3cm} \text{ if } j=2, \\ 
    \sin{X_j} \hspace{1cm} \text{ if } j=3,
    \end{cases}
\end{equation*}
\noindent $\alpha = 2$, and covariates $X_j$ drawn from an uniform distribution $U\left[-2.5, 2.5\right]$. The response follows a Gaussian distribution with $Y = \eta + \epsilon$, where $\epsilon$ is the error distributed in accordance to a $N(0, \sigma(x))$ in a homoscedastic situation with $\sigma(x) = 0.25$. Additional simulation scenarios including different response types and conditions on the variance of the error term $\epsilon$ are available at \citet{neuralGAM}. 

\begin{table}[h!]
\centering
\small
\caption{Model architecture and configuration for \CRANpkg{neuralGAM} and \CRANpkg{deepregression} in the simulated study.}
\label{tab:sim-config}
\begin{tabular}{p{2.2cm}p{5.2cm}p{5.2cm}}
\toprule
 & \textbf{\CRANpkg{neuralGAM}} & \textbf{\CRANpkg{deepregression}} \\
\midrule
\textbf{Model type} 
  & Additive neural GAM, one independent NN per predictor (backfitting + local scoring) 
  & Distributional regression with additive predictor, spline bases from \CRANpkg{mgcv} \\

\textbf{Hidden layers} 
  & 1 per term network& 1 spline basis layer\\

\textbf{Units per layer} 
  & 1024& Number of spline basis functions per smooth (thin plate regression splines, default $k=10$) \\

\textbf{Act. func.} 
  & ReLU (hidden layers), linear output 
  & Linear (basis representation) \\

\textbf{Regularization} 
  & Dropout (0.1) per hidden layer& Quadratic penalties on spline coefficients (via \CRANpkg{mgcv} penalty matrices) \\

\textbf{Optimization} 
  & Backpropagation with Adam optimizer (learning rate 0.001), trained jointly across subnetworks via backfitting 
  & Penalized likelihood optimization in TensorFlow/Keras, initialized from spline bases \\

\textbf{Uncertainty Quant.} 
  & Epistemic CIs via MC-Dropout (150 forward passes, Dropout rate $0.1$)& No dropout-based epistemic uncertainty. CIs can be obtained via model ensembles\\

\textbf{Training epochs} 
  & Maximum 10 iterations of backfitting algorithm (threshold $0.001$) & 100 epochs, early stopping with patience = 5 \\
\bottomrule
\end{tabular}
\end{table}

Both \CRANpkg{neuralGAM} and \CRANpkg{deepregression} were applied to this dataset, using smooth terms for each covariate. Performance was assessed in terms of mean squared error (MSE) on the training and test sets, and deviance explained on the training data. Table~\ref{tab:sim-results} summarizes the results, while Table~\ref{tab:sim-config} shows the architecture and configuration of each neural network. 

\begin{table}[h!]
\centering
\caption{Model performance on simulated Gaussian data.}
\label{tab:sim-results}
\begin{tabular}{lccc}
\toprule
 & MSE (Train) & MSE (Test) & Deviance Explained \\
\midrule
\CRANpkg{neuralGAM}       & 1.033 & 1.018 & 92.50\% \\
\CRANpkg{deepregression}  & 1.055 & 1.047 & 92.34\% \\
\bottomrule
\end{tabular}
\end{table}

In terms of predictive accuracy, both approaches generalize well from training to testing data, with nearly identical error rates, with \CRANpkg{neuralGAM} achieving slightly better results across all metrics. The deviance explained is above 90\% in both cases, highlighting that both methods are adequate for this problem.

\begin{figure}[h!]
    \centering
    \begin{subfigure}{0.48\textwidth}
        \centering
        \includegraphics[width=\linewidth]{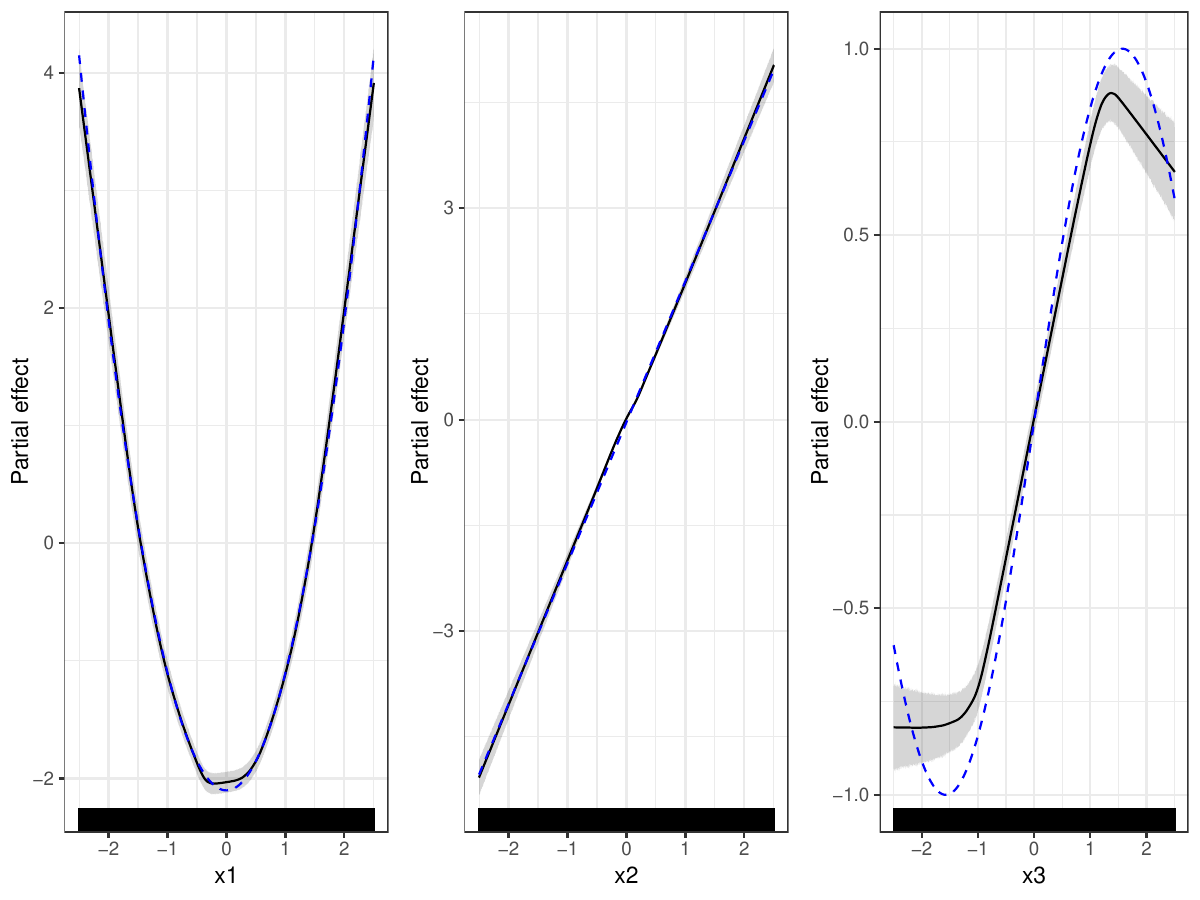}
        \caption{\CRANpkg{neuralGAM}}
        \label{fig:neural_model_plots}
    \end{subfigure}
    \hfill
    \begin{subfigure}{0.48\textwidth}
        \centering
        \includegraphics[width=\linewidth]{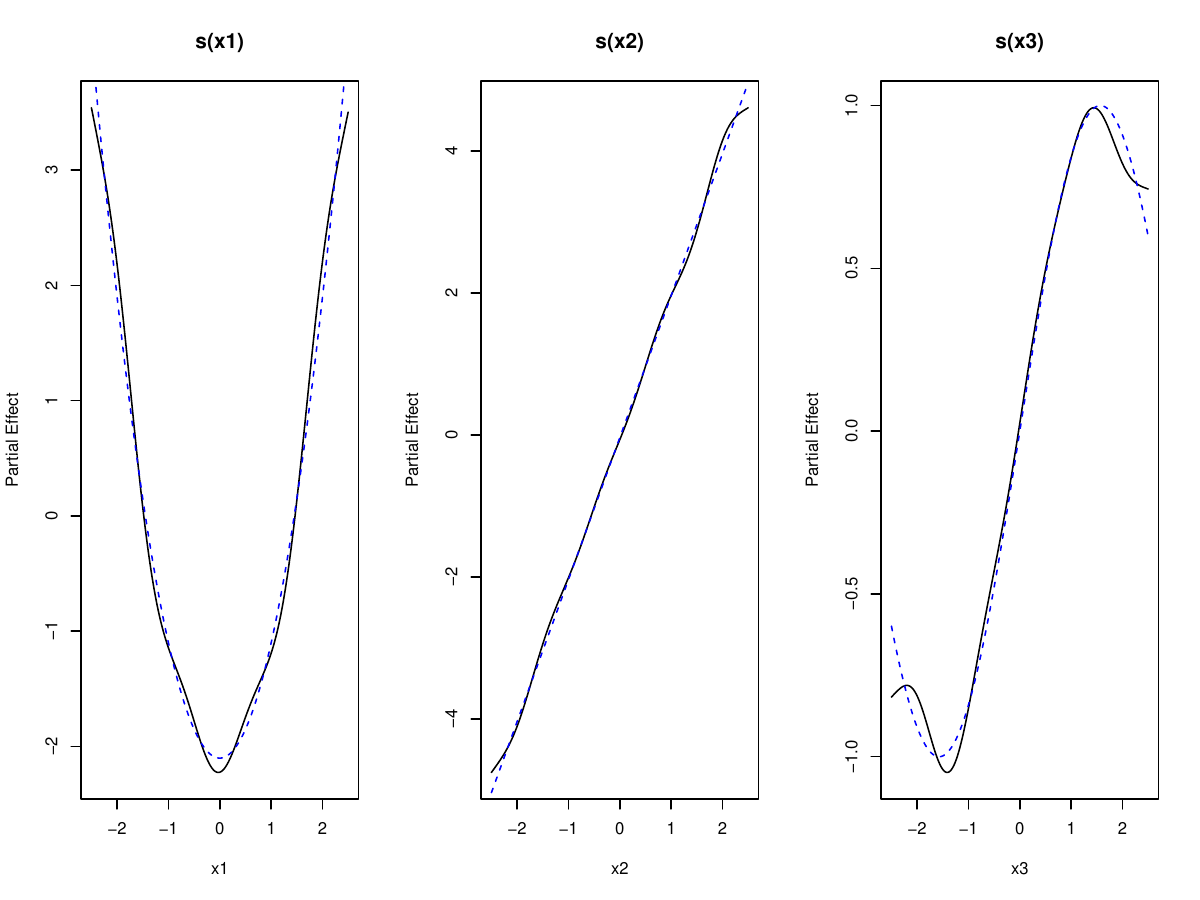}
        \caption{\CRANpkg{deepregression}}
        \label{fig:deepreg_model_plots}
    \end{subfigure}
    \caption{Estimated functions from both approaches. Black lines indicate estimated effects, blue dashed lines show the true functions. \CRANpkg{neuralGAM} includes 95\% epistemic confidence intervals obtained via MC Dropout.}
    \label{fig:compare_models}
\end{figure}

Beyond predictive metrics, we compare the obtained partial effects, comparing them to the known true functions. Figures~\ref{fig:neural_model_plots} and~\ref{fig:deepreg_model_plots} show the estimated partial effects for each covariate for \CRANpkg{neuralGAM} and \CRANpkg{deepregression}, respectively. The true shape of each function is overlaid in blue for reference. Overall, both models recover the true nonlinear effects in $X_1$ and $X_3$, as well as the linear effect in $X_2$. Both methods recover the overall patterns, but neuralGAM more closely matches the true functional form for $X_1$ and $X_2$ where \CRANpkg{deepregression} deviates more.

The \CRANpkg{neuralGAM} approach provides the added benefit of direct confidence interval estimation via epistemic uncertainty quantification using MC Dropout. Unlike \CRANpkg{neuralGAM}, \CRANpkg{deepregression} does not natively offer epistemic uncertainty through this mechanism. Instead, users must rely on model ensembles, where multiple independently trained models are combined to capture variability in the estimated functions. In this setup, ensemble predictions are obtained from the expected response conditional on a covariate grid and reference values for the other covariates. These curves reflect the full conditional mean, which includes the intercept and all other model terms, and therefore live on the outcome scale. By contrast, the plotting method for smooth terms shown in Figure\ref{fig:deepreg_model_plots} displays centered partial effects, which isolates the contribution of a single smooth and makes the scale directly comparable to the true generative functions. Consequently, ensemble uncertainty bands are not on the same vertical scale as the centered smooth effects, although the shapes of the functions remain comparable.

\begin{figure}
    \centering
    \includegraphics[width=0.5\linewidth]{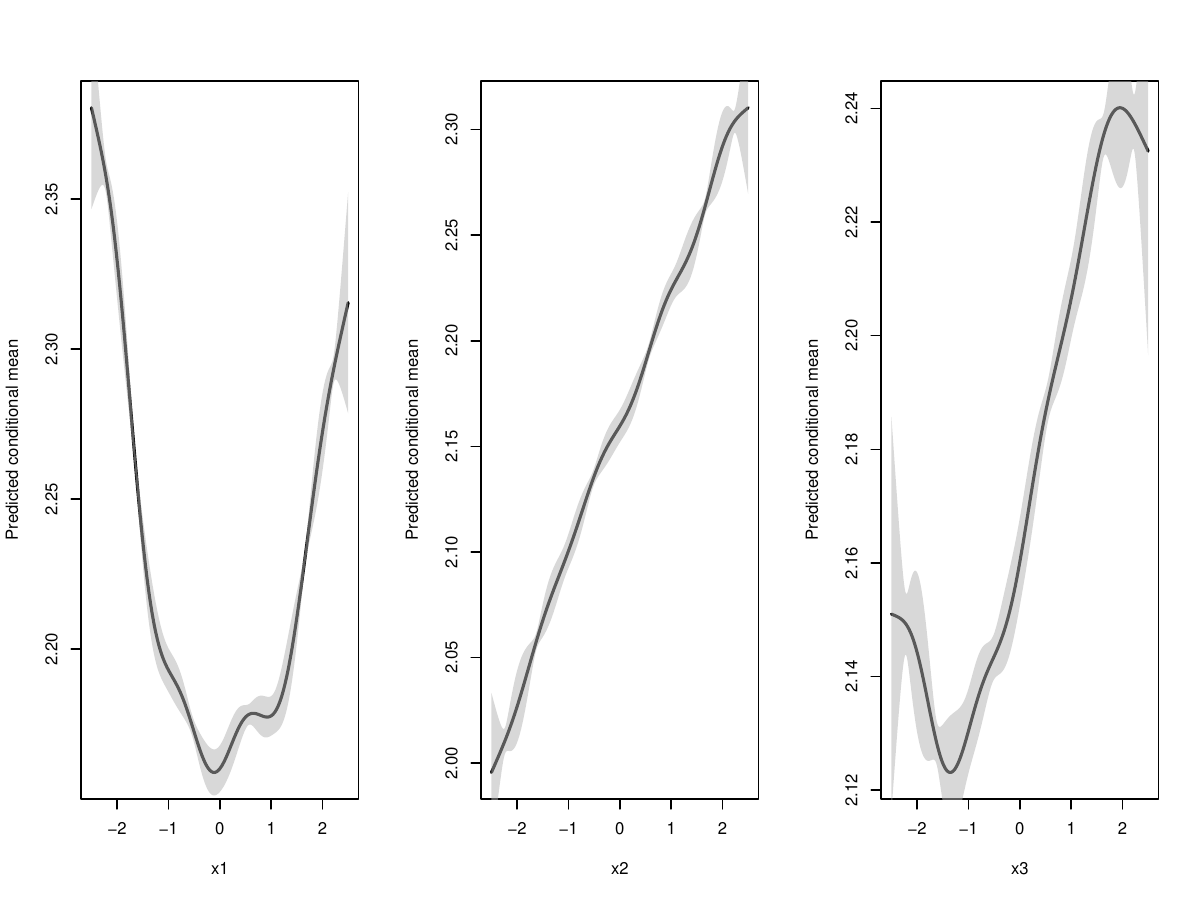}
    \caption{Estimated conditional mean functions with ensemble-based uncertainty bands from \CRANpkg{deepregression}. Black lines indicate the ensemble mean prediction, while shaded areas show epistemic uncertainty obtained from the variability across 10 independently trained models.}
    \label{fig:ensemble-deepregression}
\end{figure}

Figure~\ref{fig:ensemble-deepregression} highlights this difference in the uncertainty estimation approach: while \CRANpkg{neuralGAM} produces per-term confidence intervals via MC-Dropout (Figure~\ref{fig:neural_model_plots}), \CRANpkg{deepregression} requires ensembles to approximate epistemic uncertainty, resulting in outcome-scale bands that are not directly comparable to centered smooth effects. From a computational point of view, MC Dropout concentrates cost at \emph{inference} time, whereas ensembles multiply cost at \emph{training} time. Let $T$ be the time to train one model and $P$ the time for a single forward pass over the prediction set. With 150 MC–Dropout draws, total cost is approximately $T + 150P$ and model storage is $1\times$; with a 10-member ensemble, total cost is about $10T + 10P$ and storage is $10\times$. Since in most applications $T \gg P$, MC Dropout is typically far cheaper (one trained model, $P$ inference runs), at the expense of a larger prediction-time loop when uncertainty estimates are requested. Moreover, MC Dropout is efficient in terms of memory cost (only need to maintain one set of weights) and simplifies deployment, while ensembles can exploit hardware parallelism (training/predicting members in parallel) but require managing multiple models. In our configuration (150 draws vs.\ 10 members), MC Dropout trades $\approx150$ inference passes for avoiding $\approx9$ extra full trainings, which is usually a favorable exchange whenever predictions are not orders of magnitude more expensive than training.

While both models recover the overall quadratic, linear, and sinusoidal effects, the shapes obtained with \CRANpkg{neuralGAM} are smoother and more closely aligned with the true functions. By contrast, the \CRANpkg{deepregression} ensemble estimates show greater variability and wider uncertainty bands, especially at the boundaries, which makes their interpretation less straightforward compared to the centered partial effects provided by \CRANpkg{neuralGAM}.
\subsection{Application to real data}

We illustrate the capabilities of the \CRANpkg{neuralGAM} package with another data set. Particularly, this section details an example of its application to real data taken from the NYC Flights 13 data set from the \code{nycflights13} package \citep{nycflights13}. The data set includes airline on-time data for flights departing from all the airports in New York City during 2013. It also includes useful metadata on airlines, airports, weather conditions at different airports, and plane information.

We aim to predict whether a flight will be delayed (upon arrival) given departure flight information and certain weather conditions. With this in mind, firstly, we load the required data from the \code{nycflights13} package. We join the flights and weather data to obtain the weather conditions at a given airport (\code{origin}) and time (\code{time\_hour}). We will focus on flights departing from Newark Liberty International Airport (\code{origin == "EWR"}) in October, November, and December (\code{month \%in\% c(10,11,12)}), convert the temperature \code{temp} from Fahrenheit to Celsius, and construct a binary response variable \code{delay} which describes if the flight was delayed on arrival. The resulting class balance in the dataset is described in Table~\ref{tab:flights-balance}.

The following block of code shows how to load the required libraries and generate the EWR Oct–Dec subset and train/test splits from the \code{nycflights13} dataset:

\begin{Verbatim}
if (!require("dplyr")) install.packages("dplyr", quiet = TRUE)
if (!require("nycflights13")) install.packages("nycflights13", quiet = TRUE)

suppressMessages(library(dplyr))
suppressMessages(library(nycflights13))

seed <- 1234
set.seed(seed)

dat <- flights %>%
  filter(origin == "EWR" & month %in% c(10, 11, 12)) %>%
  left_join(weather, by = c("origin", "time_hour")) %>%
  select(arr_delay, dep_delay, air_time, temp, humid) %>%
  data.frame()

dat$temp <- (dat$temp - 32) / 1.8
dat$delay <- ifelse(dat$arr_delay > 0, 1, 0)
dat <- dat[!rowSums(is.na(dat)),]

sample <- sample(nrow(dat), 0.8 * nrow(dat))
train <- dat[sample, ]; test <- dat[-sample, ]
\end{Verbatim}

\begin{table}[h!]
\centering
\caption{Class balance summary for the EWR Oct–Dec subset.}
\label{tab:flights-balance}
\begin{tabular}{lrr}
\toprule
Class & Count & Proportion \\
\midrule
On time (0) & 16,215 & 56.8\% \\
Delayed (1) & 12,358 & 43.2\% \\
\bottomrule
\end{tabular}
\end{table}

In this scenario, we will compare three approaches: (i) \CRANpkg{neuralGAM} (binomial neural additive model with per-term subnetworks and MC–Dropout for epistemic CIs), (ii) a distributional deep regression model fitted with \CRANpkg{deepregression}, and (iii) a fully-connected, black-box neural network (Multi-Layer Perceptron, MLP) implemented in \CRANpkg{keras}. All models use the same four covariates and are evaluated by AUC–ROC on the test set. In addition, we also report the deviance explained on the training set, and perform a comparison in terms of the interpretability provided by analyzing the partial effect plots. Table~\ref{tab:sim-config-2} summarizes the architecture and hyperparameters of each method. 

\begin{table}[ht!]
\centering
\caption{Model architecture and configuration for \CRANpkg{neuralGAM}, 
\CRANpkg{deepregression}, and Keras in the simulated study.}
\label{tab:sim-config-2}

\begin{tabular}{p{2cm}p{3.6cm}p{3.6cm}p{3.6cm}}
\toprule
 & \CRANpkg{neuralGAM} 
 & \CRANpkg{deepregression} 
 & \textbf{MLP} (\CRANpkg{keras}) \\ 
\midrule

\textbf{Model type} 
& \makecell[l]{Additive neural GAM;\\ binomial family} 
& \makecell[l]{Distributional regression;\\ additive predictor;\\ spline bases from \CRANpkg{mgcv};\\ Bernoulli} 
& \makecell[l]{Feedforward MLP} \\

\textbf{Hidden layers} 
& \makecell[l]{Per-term subnetworks:\\ \textit{air\_time}: 2 layers;\\ others: 1 layer} 
& \makecell[l]{One spline-basis\\ ``layer'' per smooth} 
& \makecell[l]{2 dense\\ hidden layers} \\

\textbf{Units/layer} 
& \makecell[l]{\textit{air\_time}: (256, 128);\\ others: 128} 
& \makecell[l]{Spline basis size \\ (TPRS, default $k=10$)} 
& \makecell[l]{256, 128;\\ output: 1} \\

\textbf{Activation} 
& \makecell[l]{swish (\textit{air\_time}, \textit{temp});\\ tanh (\textit{humid})} 
& \makecell[l]{Linear\\ (basis representation)} 
& \makecell[l]{swish (hidden);\\ sigmoid (output)} \\

\textbf{Regularization} 
& \makecell[l]{Dropout $0.01$\\ per hidden layer} 
& \makecell[l]{Quadratic penalties on\\ spline coefficients} 
& \makecell[l]{No explicit weight decay;\\ early stopping} \\

\textbf{Optimization} 
& \makecell[l]{BF + LS;\\ Adam (LR $0.001$);\\ bf $10^{-2}$, ls $0.01$;\\ loss: MSE} 
& \makecell[l]{Penalized likelihood;\\TensorFlow initialized \\from spline bases} 
& \makecell[l]{Adam (LR $0.001$);\\ binary cross-entropy;\\ batch size $128$} \\

\textbf{Uncertainty} 
& \makecell[l]{MC-Dropout CIs;\\ 300 passes;\\ rate $0.01$, $\alpha=0.05$} 
& \makecell[l]{Using ensembles} 
& \makecell[l]{None} \\

\textbf{Training} 
& \makecell[l]{Iterative BF + LS \\ until convergence} 
& \makecell[l]{Up to 100 epochs;\\ early stopping\\ $\text{patience} = 5$} 
& \makecell[l]{Up to 100 epochs;\\ early stopping\\ $\text{patience} = 5$} \\

\bottomrule
\end{tabular}
\end{table}

\begin{figure}[h!]
    \centering
    \begin{subfigure}{0.48\textwidth}
        \centering
        \includegraphics[width=\linewidth]{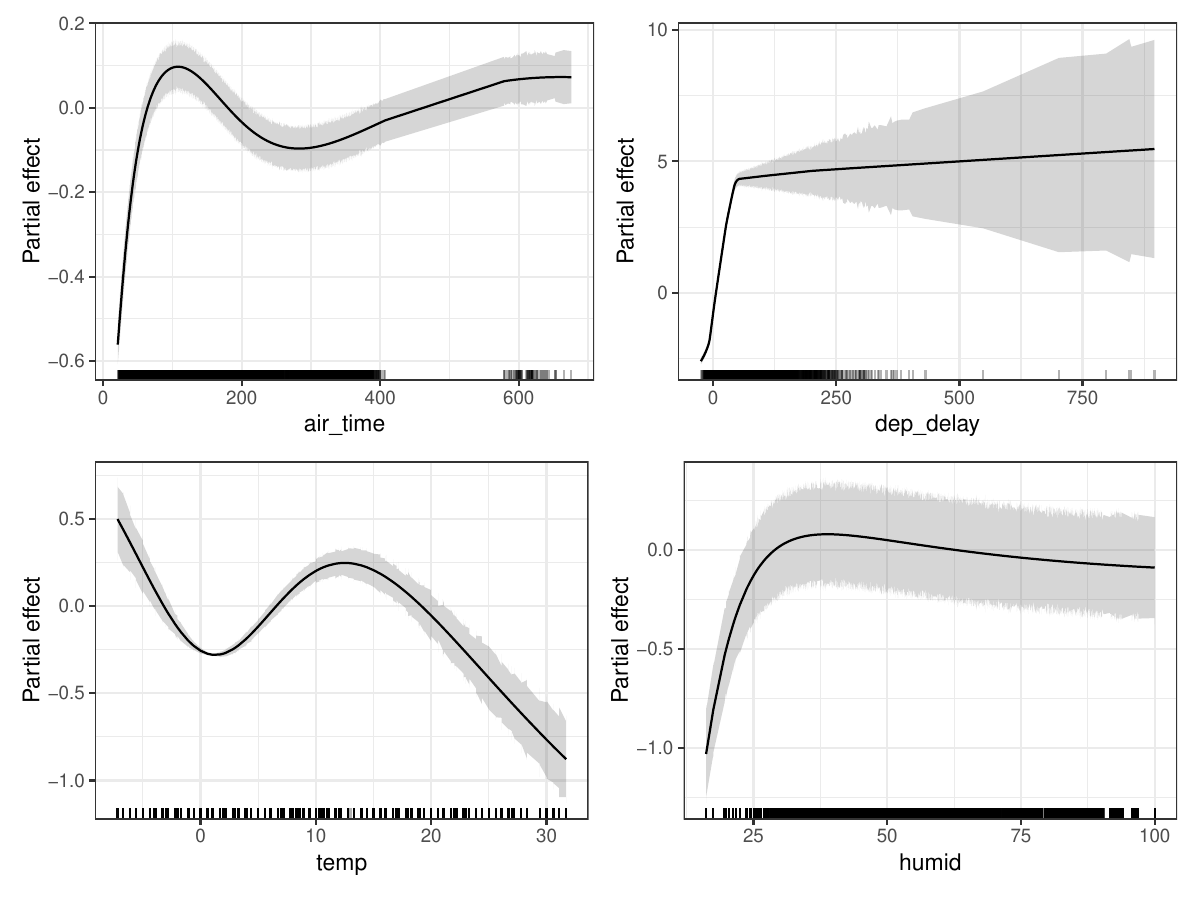}
        \caption{\CRANpkg{neuralGAM}: partial effects for each covariate with 95\% epistemic CIs obtained with \code{autoplot()}.}
        \label{fig:neural_model_flights}
    \end{subfigure}
    \hfill
    \begin{subfigure}{0.48\textwidth}
        \centering
        \includegraphics[width=\linewidth]{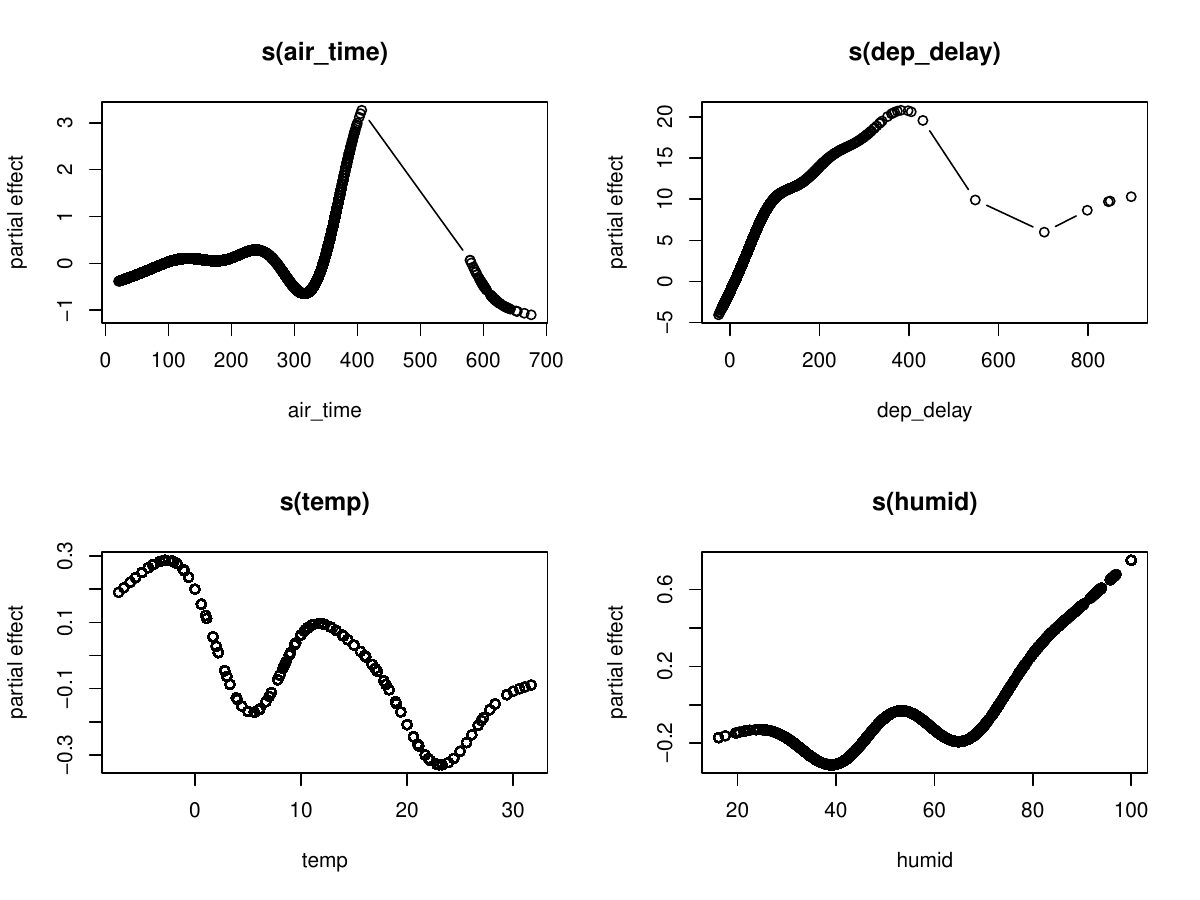}
        \caption{\CRANpkg{deepregression}: partial effects for each covariate obtained via \code{plot()}.}
        \label{fig:deepreg_airtime_binomial}
    \end{subfigure}
    \caption{Estimated partial effects for the four covariates.}
    \label{fig:comparison_flights}
\end{figure}

In this case, we showcase how \CRANpkg{neuralGAM} can be personalized to use custom neural network architectures for each covariate for greater flexibility. For instance, the integration of \CRANpkg{neuralGAM} with \CRANpkg{keras} enables the direct use of different activation and loss functions. In this case, we showcase how to use \emph{Swish} \citep{ramachandran2017searching}, a self-regularization, non-monotonic activation function that has demonstrated stronger performance than ReLu across a number of challenging datasets, to model the \code{air\_time} and \code{temp} covariates:

\begin{Verbatim}
ngam_bin <- neuralGAM(
  delay ~ s(air_time, activation = "swish", num_units = c(256,128)) +
    s(dep_delay) +
    s(temp, activation = "swish") +
    s(humid, activation = "tanh" ),
  data = train, family = "binomial",
  num_units = 128, loss = "mse",
  bf_threshold = 1e-2, ls_threshold = 0.01, alpha = 0.05,
  uncertainty_method = "epistemic", forward_passes = 300,
  dropout_rate = 0.01,
  seed = seed, verbose = 1
)
\end{Verbatim}

In the case of \CRANpkg{deepregression}, we specify smooth functions for all covariates to model the mean distribution and \code{family = "bernoulli"} to indicate a logistic regression model. We fit the model for 100 epochs (with early stopping), and compute the MSE and deviance explained by the model: 

\begin{Verbatim}
deepreg_mod_bin <- deepregression::deepregression(
  y = train$delay, data = train,
  list_of_formulas = list(
    loc = ~ 1 + s(air_time) + s(dep_delay) + s(temp) + s(humid)
  ),
  list_of_deep_models = NULL, family = "bernoulli"
)

deepreg_mod_bin %>% fit(
  epochs = 100, view_metrics = FALSE, early_stopping = TRUE, patience = 5
)
\end{Verbatim}

At last, to compare the performance of the neural-network based GAMs with a classic, black-box deep neural network, we fit an analogous neural network model using \CRANpkg{keras}:

\begin{Verbatim}
x_train <- as.matrix(train[, c("air_time","dep_delay","temp","humid")])
y_train <- train$delay
x_test  <- as.matrix(test[, c("air_time","dep_delay","temp","humid")])
y_test  <- test$delay

keras_model <- keras_model_sequential() %>%
               layer_dense(units = 256, activation = "swish", input_shape = ncol(x_train)) %>%
               layer_dense(units = 128, activation = "swish") %>%
               layer_dense(units = 1, activation = "sigmoid")

keras_model %>% compile(
  optimizer = optimizer_adam(learning_rate = 0.001),
  loss = "binary_crossentropy", metrics = c("accuracy")
)

keras_model %>% fit(x_train, 
                    y_train, 
                    epochs = 100, 
                    batch_size = 128, 
                    verbose = 1, 
                    shuffle = FALSE,
                    view_metrics = FALSE, 
                    callbacks = list(callback_early_stopping(monitor = "accuracy", patience = 5)))
\end{Verbatim}

\begin{table}[ht!]
\centering
\caption{Predictive performance on the flight delay task.}
\label{tab:comparison_realdata}
\begin{tabular}{lcc}
\toprule
Model & AUC–ROC (Test)  & Deviance Explained (Train) \\
\toprule
\CRANpkg{neuralGAM}    & 0.819 & 30.89\% \\
\CRANpkg{deepregression}    & 0.838 & 34.09\% \\
\CRANpkg{keras} & 0.838 & 34.02\% \\
\bottomrule
\end{tabular}
\end{table}

Table~\ref{tab:comparison_realdata} shows the obtained MSE in the test set and the deviance explained by each of the trained models. We can observe how the deviance explained by \CRANpkg{deepregression} ($34.09\%$) is similar to that obtained by \CRANpkg{neuralGAM} ($30.89\%$). Regarding performance on the test set, the AUC-ROC achieved by \CRANpkg{deepregression} was $0.838$, slightly higher than the value of $0.817$ achieved by \CRANpkg{neuralGAM} and comparable to the black-box neural network ($0.830$). 

Black-box NNs can provide slightly higher predictive performance, achieving an AUC-ROC of $0.838$ but offer no interpretability regarding the contribution of individual covariates. In contrast, \CRANpkg{neuralGAM} and \CRANpkg{deepregression} achieve comparable performance while enabling visualization of partial effects and uncertainty estimates. The black-box neural network, trained with two dense swish layers and a sigmoid output, learns a fully non-additive mapping between predictors and the outcome. This unconstrained approach allows it to capture interactions that additive models inherently miss and can sometimes improve predictive accuracy. However, it lacks the interpretability of \CRANpkg{neuralGAM} or \CRANpkg{deepregression}: no individual partial effects can be directly visualized, and it provides no intrinsic uncertainty quantification.

At last, focusing on the interpretability provided by the GAM-style methods, Figure~\ref{fig:comparison_flights} compares the estimated partial effects of \CRANpkg{neuralGAM} and \CRANpkg{deepregression} for the four covariates in the flight delay model. Again, it is worth mentioning that \CRANpkg{deepregression} does not provide CI estimates directly, and must be obtained through computationally expensive model ensembles.

First, focusing on the partial effect of \code{air\_time} \CRANpkg{neuralGAM} shows a smooth non-monotonic, non-linear pattern that increases steeply for short flights (up to about 100 minutes), reaches a modest peak, and then gradually decreases for longer flights up to 400 minutes, and the delay probability increases again for flights with air time $>600$ minutes. The model captures this relationship with smooth transitions and stable uncertainty, reflected by the relatively narrow confidence bands even at the extreme tail (beyond 600 minutes). In contrast, \CRANpkg{deepregression} displays a more irregular curve for \code{s(air\_time)} with sharper oscillations: the shape fluctuates more abruptly, suggesting that the spline-based estimation may be more sensitive to local variations or noise. While this can offer flexibility, it comes at the cost of interpretability and stability, particularly in regions with limited data.

In the case of \code{dep\_delay}, both models agree that higher departure delays increase the probability of arrival delay.
For \CRANpkg{neuralGAM}, the effect rises sharply up to approximately 100 minutes of departure delay and then flattens, showing a saturation effect. The narrow confidence interval in the main data region (0–300 minutes) demonstrates strong certainty and robust estimation, but the band widens beyond $\sim 300$ minutes, where data become sparse. In the case of \CRANpkg{deepregression}, the effect of \code{dep\_delay} appears almost linear across the range, with a continuous increase up to 400 min.

The temperature effect estimated by \CRANpkg{neuralGAM} shows a smooth, nonlinear pattern in which flight delays are most likely at very cold temperatures (around –5ºC), then decrease sharply as conditions warm. The delay probability rises again between 5 °C and 10 °C, possibly reflecting transitional weather effects such as fog or humidity, and then declines steadily beyond 15ºC, indicating that moderate temperatures are generally most favorable for on-time departures. Overall, the curve captures a complex relationship where both extremely low and moderately cool temperatures elevate delay risk, while warm, stable conditions correspond to smoother flight operations. The confidence interval around the estimated curve remains relatively narrow across most of the temperature range, implying that the model’s uncertainty is low and the estimated effect is stable. Only at the extreme ends of the temperature distribution the interval is wider, which reflects data sparsity in those regions. Meanwhile, \CRANpkg{deepregression} learns a more wiggly pattern with several local extrema across the range, including more oscillations between 10–30ºC. These small fluctuations suggest overfitting to local noise, particularly where data density is lower.

At last, for the effect of \code{humid}, both models suggest that higher humidity implies a higher probability of delay.
In \CRANpkg{neuralGAM}, this pattern is monotonic and smooth, with an initially steep rise that levels off beyond 50\% humidity. The narrow confidence bands indicate a stable and well-regularized estimate. By contrast, \CRANpkg{deepregression} shows a wigglier response, with alternating increases and decreases before ultimately trending upward.

Overall, the \CRANpkg{neuralGAM} results exhibit smooth, stable, and interpretable effects due to its neural network based regularization. Moreover, the Monte Carlo Dropout based intervals are coherent and narrow in data-dense regions, indicating reliable uncertainty quantification. In contrast, \CRANpkg{deepregression}, while grounded in the spline framework from \CRANpkg{mgcv}, tends to produce more irregular and wiggly smooths, especially in sparse data regions. This flexibility allows it to capture more complex shapes but also makes it more susceptible to overfitting and harder to interpret.

\section{Summary and discussion} \label{sec:summary}
In this paper, we introduced \CRANpkg{neuralGAM}, an \textsf{R} package for fitting Generalized Additive Neural Networks (GANNs) which combines the interpretability of additive models with the flexibility of neural networks. The proposed methodology extends the classical Generalized Additive Model (GAM) framework by using independent neural networks to estimate the contribution of each covariate to the response, ensuring additivity through the local scoring and backfitting algorithms. This white-box modeling approach enables the visualization and interpretation of each feature's partial effect while maintaining the ability to learn complex, non-linear patterns from data.

To contextualize the contribution of \CRANpkg{neuralGAM}, we compare it against \CRANpkg{deepregression}, a package that embeds \CRANpkg{mgcv}-style spline bases and associated penalty matrices into deep learning architectures. Both approaches combine additive modeling with neural networks, but they differ substantially in scope and estimation strategy. \CRANpkg{deepregression} is designed for distributional regression, allowing simultaneous modeling of mean, scale, and shape parameters, with smooth terms regularized through penalized likelihood optimization in TensorFlow/Keras. In contrast, \CRANpkg{neuralGAM} focuses on interpretable additive predictors for the mean response, fitting one independent neural network per predictor via local scoring and backfitting.

Through comprehensive examples on both simulated and real data, we demonstrated the ability of \CRANpkg{neuralGAM} to recover smooth effect estimates comparable to those obtained using other methods such as \CRANpkg{deepregression} and generalize well in predictive tasks, while offering high flexibility, particularly in modeling sharp nonlinearities and capturing complex patterns in sparse regions. Importantly, \CRANpkg{neuralGAM} is, to the best of our knowledge, the only \textsf{R}-based implementation of a GANN using independently-fitted (deep) neural networks, positioning it as a valuable tool for researchers seeking interpretable neural network implementations.

Overall, \CRANpkg{neuralGAM} offers a strong compromise between interpretability and the flexibility provided by neural network-based estimation, producing stable, smooth, and easily interpretable effect estimates with uncertainty quantification. \CRANpkg{deepregression} remains a valuable benchmark for highly flexible spline-based additive modeling, but our initial experiments show that it can be more prone to overfitting and less stable in sparse regions. The Keras black-box neural network represents the opposite end of the spectrum: a powerful, unconstrained model focused on predictive performance but with limited transparency. These three approaches illustrate the trade-offs between interpretability, flexibility, and predictive power that need to be balanced in modern statistical learning.

The current version of the package supports Gaussian, binomial, and Poisson response distributions, making it suitable for multiple applications. As future work, several research directions have been identified to further extend the proposed algorithm. A first venue concerns the generalization of the model to allow multinomial logistic regression, thereby enabling its application to multi-class classification problems. 

A second line of extension concerns the incorporation of interaction terms, allowing the model to capture joint effects between covariates. 
The local scoring algorithm can be adapted to estimate such terms, enabling the model to learn not only the marginal contributions of individual predictors but also their potential interactions.

At last, we also foresee the incorporation of support for automatic feature selection procedures, support for cyclic and spatial features, and the inclusion of additional uncertainty estimation methods, including conformal prediction \citep{karimi2023quantifying}, bootstrap-based methods, or the use of empirical quantiles to obtain confidence intervals from MC Dropout.

In terms of future software-development improvements, we foresee the support for additional deep learning backends and APIs beyond Keras and Tensorflow. 

These additions will further enhance the interpretability, flexibility, and applicability of the model across a broader range of statistical and machine learning problems.

\section*{Computational details}

The results in this paper were obtained using \textsf{R}~4.4.1 (2024-06-14 ucrt) with the \CRANpkg{neuralGAM}~2.0.1 package. \textsf{R} itself
and all packages used are available from the \href{https://CRAN.R-project.org/}{Comprehensive \textsf{R} Archive Network (CRAN)}.

\section*{Acknowledgments}
This work was supported by the project ``FAIR'', financed by the ``European Union NextGeneration-EU'' (\url{https://next-generation-eu.europa. eu/index_es}), the Recovery Plan, Transformation and Resilience (\url{https://planderecuperacion.gob.es/}), through Spanish National Cybersecurity Institute (INCIBE) (\url{https://www.incibe.es/})” program under the fourth CPI call (CPI-2023) and the Xunta de Galicia (Centro singular de investigaci\'{o}n de Galicia accreditation 2019-2022) and the European Union (European Regional Development Fund - ERDF); and the Grant PID2023-148811NB-I00 (MINECO/AEI/FEDER, UE).

\bibliography{filename}

\end{document}